\documentclass[final,3p]{elsarticle}

\usepackage[colorlinks=true,breaklinks=true,pdftex]{hyperref}
\usepackage{pifont}
\usepackage{wrapfig}
\usepackage{lipsum}
\usepackage{amsmath}
\usepackage{amssymb}
\usepackage{booktabs}
\usepackage{wrapfig}
\usepackage{lipsum}
\usepackage{multirow}
\usepackage{pifont}
\usepackage{float}
\usepackage{url}
\usepackage{latexsym}
\usepackage{color}
\usepackage{xcolor}

\usepackage{lipsum}  
\usepackage{afterpage}

\definecolor{newcolor}{rgb}{.8,.349,.1}
\renewcommand{\textcolor}[2]{#2}
\journal{GMP 2024}

\newcommand{\qs}[1]{\textcolor{black}{{#1}}}

\biboptions{authoryear}

\begin{document}

\begin{frontmatter}


\title{TaylorGrid: Towards Fast and High-Quality Implicit Field Learning via Direct Taylor-based Grid Optimization}

\fntext[fn1]{Equal contribution.} 
\cortext[cor1]{Corresponding author.}
\author[1]{Renyi Mao \fnref{fn1}}
\emailauthor{maory21@emails.jlu.edu.cn}{Renyi Mao}
\author[2]{Qingshan Xu \fnref{fn1}}
\emailauthor{qingshan.xu@ntu.edu.sg}{Qingshan Xu}
\author[1]{Peng Zheng }
\emailauthor{zhengpeng22@mails.jlu.edu.cn} {Peng Zheng}
\author[1]{Ye Wang }
\emailauthor{yewang22@mails.jlu.edu.cn}{Ye Wang}
\author[1,3]{Tieru Wu}
\emailauthor{wutr@jlu.edu.cn}{Tieru Wu}
\author[1,3]{Rui Ma \corref{cor1}}
\emailauthor{ruim@jlu.edu.cn}{Rui Ma}

\address[1]{School of Artificial Intelligence, Jilin University, Changchun, 130012, China}
\address[2]{School of Computer Science and Engineering, Nanyang Technological University, Singapore, 639798, Singapore}
\address[3]{Engineering Research Center of Knowledge-Driven Human-Machine Intelligence, MOE, China}



\begin{abstract}


Coordinate-based neural implicit representation or implicit fields have been widely studied for 3D geometry representation or novel view synthesis.
Recently, a series of efforts have been devoted to accelerating the speed and improving the quality of the coordinate-based implicit field learning.
Instead of learning heavy MLPs to predict the neural implicit values for the query coordinates, neural voxels or grids combined with shallow MLPs have been proposed to achieve high-quality implicit field learning with reduced optimization time.
On the other hand, lightweight field representations such as linear grid have been proposed to further improve the learning speed.
In this paper, we aim for both fast and high-quality implicit field learning, and propose TaylorGrid, a novel implicit field representation which can be efficiently computed via direct Taylor expansion optimization on 2D or 3D grids.
As a general representation, TaylorGrid can be adapted to different implicit fields learning tasks such as SDF learning or NeRF. 
From extensive quantitative and qualitative comparisons, TaylorGrid achieves a balance between the linear grid and neural voxels, showing its superiority in fast and high-quality implicit field learning.

\end{abstract}


\end{frontmatter}


\section{Introduction}
Coordinate-based implicit fields represent complex and continuous signals in a compact way. Therefore, they have attracted great interests in computer vision and graphics, such as 3D shape reconstruction \cite{park2019deepsdf,mescheder2019occupancy,chen2019learning,fu2022geo}, scene completion \cite{sitzmann2020implicit}, novel view synthesis \cite{BenMildenhall2020NeRFRS,barron2021mip,yariv2021volume} etc. In general, coordinate-based implicit fields usually encode or parameterize target signals by a large multi-layer perceptron (MLP), \textit{e.g.}, DeepSDF \cite{park2019deepsdf} for shape reconstruction and Neural Radiance Fields (NeRFs) \cite{BenMildenhall2020NeRFRS} for novel view synthesis. However, this representation usually requires a lengthy training time ranging from hours to days for a single scene, which greatly limits the real-world applications of coordinate-based implicit fields.

\begin{figure*}[t]
\centering
\includegraphics[width=0.9\columnwidth]{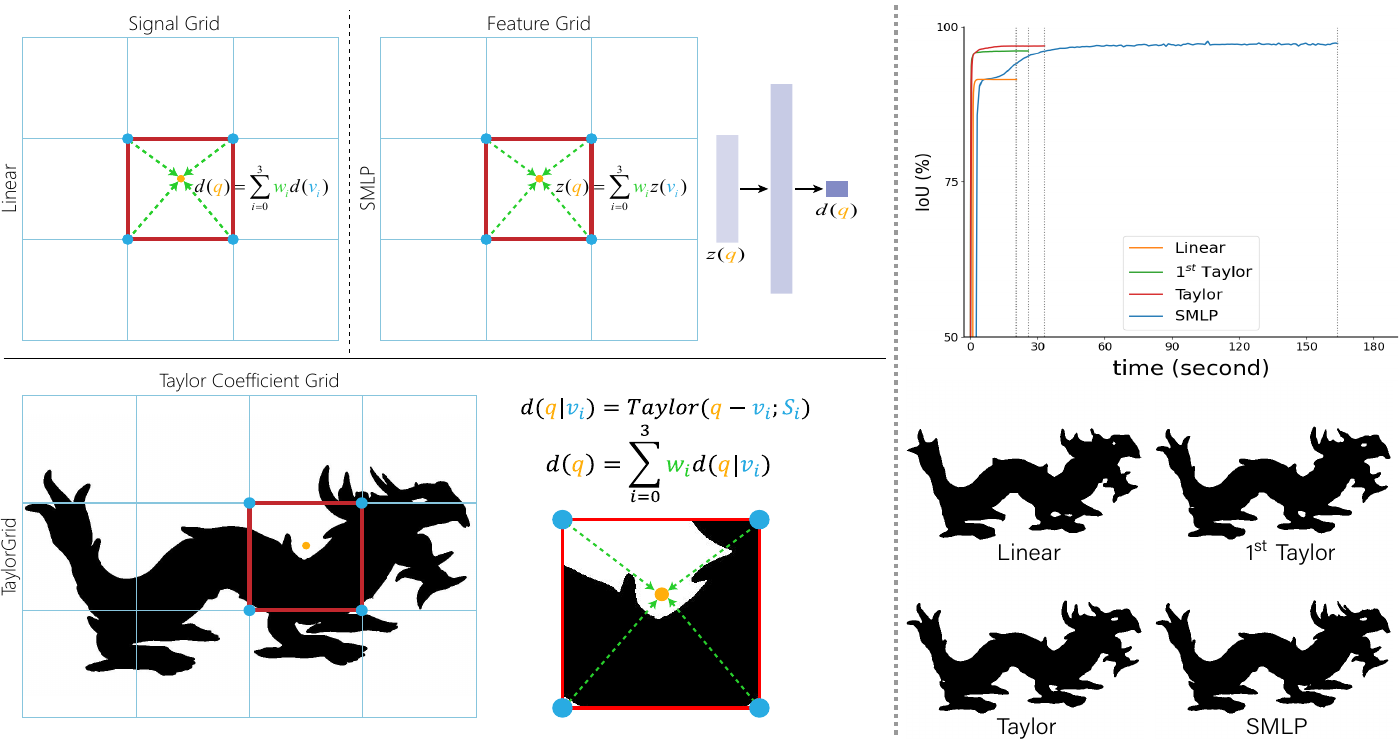}
  \caption{
    Linear gird methods store a signal $d\left(\textcolor{cyan}{v_i}\right)$ at each grid vertex, and get the query result with tri-linear interpolation. SMLP methods store features $z\left(\textcolor{cyan}{v_i}\right)$ at grid vertices, and first get the feature of a queried point, $z\left(\textcolor{orange}{q}\right)$, by tri-linear interpolation. Then, SMLP methods decode the feature into an implicit value with a \textcolor{violet!70}{shallow MLP}. We use Taylor expansion series to fit the signals of an implicit field. Our method is simple yet effective, and it stores a group of Taylor expansion coefficients at each grid vertex, which can be used to approximate the implicit value with Taylor polynomial formulation. Then, we combine the approximated values computed with all neighboring grid vertices with tri-linear interpolation. Our method is efficient 
    and effective, 
    with a convergence rate comparable to linear grid methods and a quality comparable to shallow MLP methods.
    }
  \label{pre_all}
\end{figure*}

For any queried point, the coordinate-based MLPs need to go through the whole network to learn its target signal. Therefore, coordinate-based MLPs are computationally expensive for implicit fields. In contrast, methods based on the linear grid incorporate the linear signal grids with some non-linearity \cite{karnewar2022relu} to directly model the target signals (cf. the top left of Fig.~\ref{pre_all}). For a queried point, these methods only need to search for its spatial grid neighbors to compute the target property such as signed distance function (SDF) or density value. Therefore, the implicit methods based on linear grids can achieve a superior speed. However, the representation capability of signal grids is limited, which usually leads to the performance degradation.
On the other hand, by combining the advantages of MLPs and signal grids, neural voxel methods such as \cite{voxGo} employ a feature grid together with a feature-conditioned shallow MLP to model the target property, denoted as SMLP (shallow MLP) methods in this paper (cf. the top middle of Fig.~\ref{pre_all}).
These SMLP methods can take both speed and quality into consideration. 
However, due to the usage of shallow MLPs, their efficiency is still much worse than that of linear grid methods.
Then, a motivating question is, whether it is possible and how to directly use the grids to model the target property so that both speed and quality are guaranteed at the same time.

In this paper, we propose a novel implicit field representation that employs low-order Taylor expansion formulations to optimize the grids for encoding the field signals, such as volume density and SDF. Specifically, the data stored at each grid vertex are  the coefficients of low-order Taylor expansion which approximates the signal functions. The implicit value of a queried point is computed based on the grid values (Taylor coefficients) of its spatial grid neighbors.
This new representation, coined as TaylorGrid, has several strengths. First, our Taylor-based grid representation is endowed with more powerful representation capabilities due to the additional continuous non-linearity introduced by the low-order Taylor expansion formulations. This guarantees good quality of our results in applications such as geometry reconstruction and novel view synthesis. Second, without any neural network, our TaylorGrid is more compact comparing to the SMLP methods and is efficient in terms of both time and memory. Third, our approach can be easily integrated into existing methods which require encoding the implicit fields and predicting the queried values. This allows our TaylorGrid to be widely applicable to different implicit field learning tasks.

\begin{figure}[t]
\centering
\includegraphics[width=0.5\columnwidth]{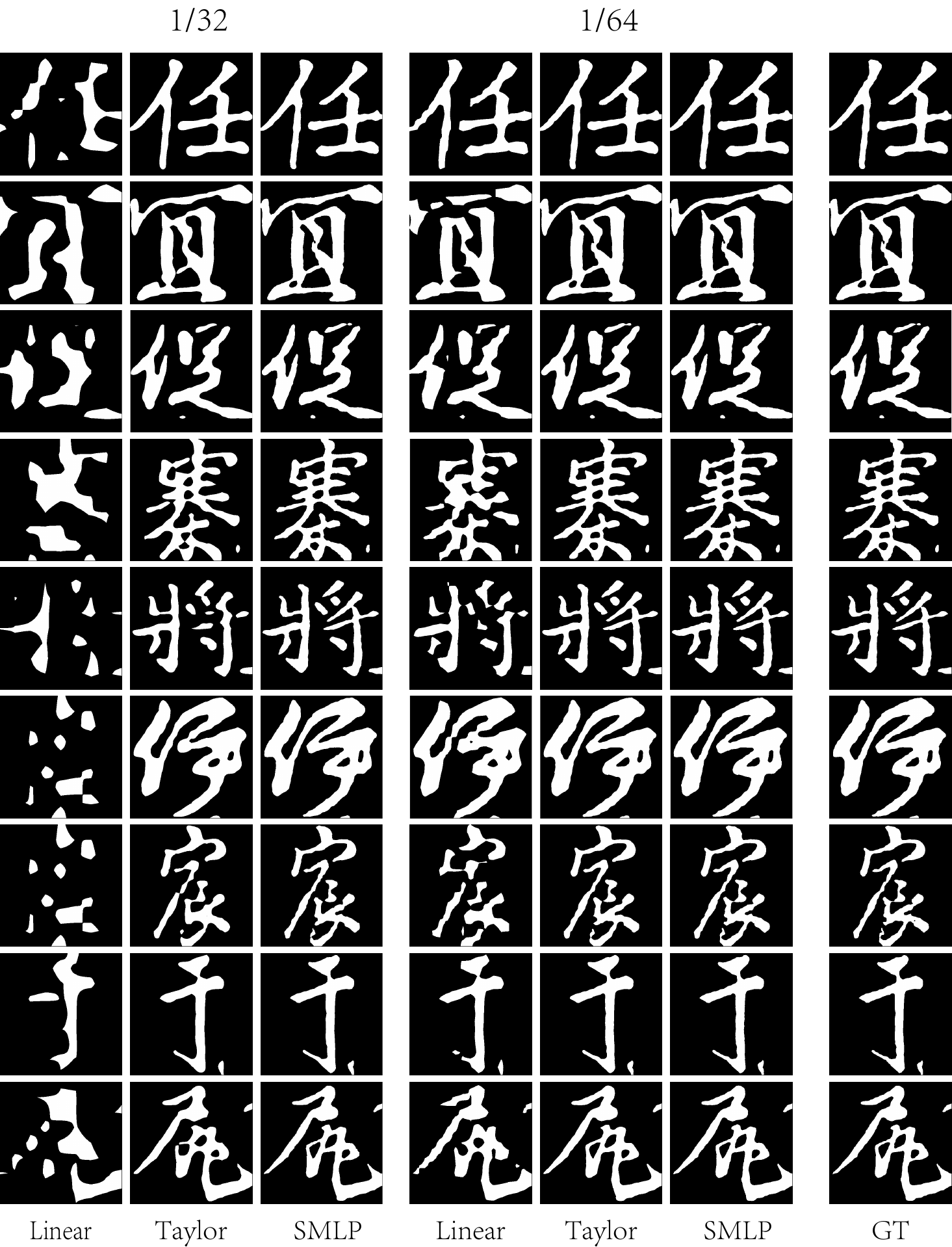}
  \caption{We use the Kuzushiji-MNIST \cite{kMNIST} handwritten dataset to reconstruct image-based signed distance fields (SDF), and demonstrate the reconstruction results of different methods at different grid resolutions. The $1/64$ represents the grid resolution is 64 times smaller than the original image while $1/32$ represents the grid resolution is 32 times smaller than than the original image. 
  Each column shows the SDF reconstruction result of images, as well as the ground truth images on the rightmost column. 
  We use methods such as linear interpolation, Taylor grid, neural voxel combined with a small MLP. 
  }
  \label{2D_fig}
\end{figure}

In summary, our contributions are as follows: 1) We propose TaylorGrid, a novel implicit filed representation which can be efficiently computed via direct Taylor expansion optimization on 2D or 3D grids. This representation bridges the gap between linear grids and neural voxels. 2) We show that the optimization of TaylorGrid can converge and become stable quickly like linear grids, while encoding complex field signals like neural voxels, as shown in the bottom right of Fig.~\ref{pre_all} and Fig.~\ref{2D_fig}. 3) We have demonstrated the flexibility and practicality of TaylorGrid by replacing the field representation in different models. Extensive quantitative and qualitative comparisons show our representation achieves a better trade-off between efficiency and representation capability against other representations for tasks such as 3D reconstruction and novel view synthesis.


\section{Related Work}
\subsection{Implicit Field Representation}
\par In computer graphics, object geometry can be represented in various ways, such as mesh, point cloud, and implicit representation. With the prosperity of deep learning, implicit representation has received wide attention. DeepSDF \cite{park2019deepsdf}, IM-Net \cite{chen2019learning} and Occupancy-Net \cite{mescheder2019occupancy} propose deep implicit presentations for geometry. All these methods represent the geometric surface of objects as level-sets and use cumbersome MLPs to predict the implicit values. And then the mesh can be extracted by the marching cube \cite{marchcube}. BSP-Net \cite{chen2020bsp} combines space binary partition tree with neural networks, and realizes a compact and interpretable implicit representation. CVXNET \cite{deng2020cvxnet} also puts forward a similar work. 
PRIF \cite{feng2022prif} implements an implicit rendering-oriented representation for ray-based distance modeled with Plucker coordinates. 
Deep marching tetrahedra \cite{tetrahedra} combines tetrahedron with SDF. By inheriting the advantage of the mesh representation, this semi-implicit representation can be easily used for rendering and deformation. 
Besides the geometric implicit fields, NeRF \cite{BenMildenhall2020NeRFRS} represents scenes as a density field and a color field, advancing the field of novel view synthesis. The rendering-oriented density field can be transformed from SDF under the efforts of NeuS \cite{wang2021neus} and VolSDF \cite{yariv2021volume}.

\subsection{Speedup of Implicit Field Learning}
\par Neural representation methods proposed by DeepSDF \cite{park2019deepsdf} and NeRF \cite{BenMildenhall2020NeRFRS} have made a progress in computer vision and graphics. However, because they use coordinate-based MLPs, their representation power is limited and the computational cost is expensive. 
In order to improve the representation power, LIG \cite{genova2020LIG} and deep local shape
\cite{chabra2020deeplocal} exploit feature voxels to encode input information and use MLPs to decode information. 
This semi-implicit representation greatly improves the expression ability and efficiency of the model, but the training speed is still slow with cumbersome MLPs. 

NGLOD \cite{Nglod}, Acron \cite{acron}, DualOctree \cite{wang2022dual}, and INSTANT-NGP \cite{muller2022instant} encode the scene information into sparse data structures, which greatly improves the speed. 
NGLOD first improves the rendering speed of neural surface SDF to real-time rendering level with octree-based multi-level of details. 
For radiance field, as the ray sampling would induce millions of point queries for the neural network, designing sparse structure in radiance field is also important for speeding up training and test. 
NVSF \cite{NVSF} proposes to build a sparse voxel field during training. Although the training speed is greatly improved, it is still slow due to use of cumbersome MLPs.
PlenOctree \cite{yu2021plenoctrees} first learns the entire scene and builds an Octree to achieve real-time rendering of NeRF after well training, but the training still takes considerable time. 
 TensoRF \cite{chen2022tensorf} uses the tensor decomposition method to solve the problem of slow high-dimensional grid interpolation, thus improving the speed. In order to quickly build sparse data or avoid points sampling on empty space during training, DVGO \cite{voxGo}, Plenoxels \cite{fridovich2022plenoxels} and VoxGRAF \cite{VoxGRAF} use linear grid to represent the density field, while this explicit geometry representation makes it convenient to determine whether a certain region is empty, and reduces the points sampling in empty regions. 
Although the linear grid converges fast, its representation power is limited. 
Relu Field \cite{karnewar2022relu} proposes to introduce non-linearity with activation functions defined on linear grid to improve its expressive power. 
Our method does not focus on designing a sparse data structure to improve speed,
but targets to design a general and improved representation by introducing non-linearity with Taylor formulation, which retains the convergence characteristics of linear grid but has higher quality.

\subsection{Taylor-based Representation}
\par TaylorIMNet \cite{xiao2022taylorimnet} proposes to predict the implicit values at query points using Taylor series computed at nearby expansion points, so as to avoid the use of cumbersome MLPs in test, but this does not help the training efficiency since MLPs still need to be trained to predict Taylor coefficients. 
Inspired by this work, we propose to directly use Taylor-based grid to encode the entire scene, instead of using some network to predict the Taylor series coefficients at any point in the scene. 
Specifically, we directly anchor the Taylor expansion points on the grid vertices, and perform the scene fitting and prediction \textit{without any network}. TaylorNet \cite{zhaotaylornet} proposes to use a single high-order Taylor series as the parameters of neural network without nonlinear activation function, and uses tensor decomposition to solve the problem of exponential growing for high-order Taylor series.
Gradient-SDF \cite{gradientSDF} proposes to stores the SDF values and its gradients at each voxel (which can also be treated as a 1-st order Taylor-based representation), so it can efficiently find the zero level-set of SDF and is well used to depth tracking.
For our method, we observed the representation power increases slowly and the coefficients increase exponentially as the order of the Taylor series increases. Thus, we propose to use low-order Taylor series to fit the implicit signals of the target scene.
Also, we propose an efficient pipeline to direct optimize the Taylor-based grid.

\section{Method}
\qs{In this work, our goal is to design a pure grid-based representation with Taylor expansion formulation to encode the implicit field signals such as SDF or density. In this way, our method can take both efficiency and representation power into consideration. The pipeline of our method is shown in Fig.~\ref{pre_all}, which demonstrates how the TaylorGrid is defined to compute the signed distance or density at a query point. In the following, we first introduce our method in Section~\ref{Taylor expansion formulation}. Then, Section~\ref{gemotry} presents the application of TaylorGrid to 3D geometry reconstruction. At last, we further present the application of our method to novel view synthesis.}

\subsection{TaylorGrid} \label{Taylor expansion formulation}
\par \qs{Given an objective} function encoding the geometry $\emph{ f } : \mathbb{R}^{n} \rightarrow \mathbb{R}$, in its variable space, 
\qs{for a point $x \in \mathbb{R}^{n}$, Taylor expansion expresses the relationship between its function value $\emph{ f }\left( x \right)$ and the derivatives of its neighboring expansion point $ x_p \in \mathbb{R}^{n} $.}
We \qs{denote} the $\emph{n}$-order  derivative of $\emph{ f } $ \qs{at} $x_p$ as $\emph{ f }^{ \left(n\right)} \left( x_p \right)$. 
\qs{In this way, we can approximate $\emph{ f }\left( x \right)$ by  $\emph{ T } \left( x-x_p;  S(x_p) \right)$ as follows:}
\begin{equation} \label{Taylor_expansion}
        \emph{ T } \left( x-x_p ;  S(x_p) \right)
        = \emph{ f }^{0} \left(x_p\right)+ \emph{ f }^{1} \left( x_p \right)^{\qs{\top}}\left( x - x_p \right) + \frac{1}{2!}\left( x - x_p \right)^{\qs{\top}}  \emph{ f }^{2} \left( x_p \right)\left( x - x_p \right) +  \dots,
\end{equation}
\qs{where $S(x_p) = \{f^0(x_p), f^1(x_p), f^2(x_p), \dots\}$ represents Taylor series, which relies on the map function $f$ and the expansion point $x_p$.}
\qs{Commonly, $f^1(x_p)$ and $f^2(x_p)$ are denoted as $\Delta\emph{f}$ and $\nabla\emph{f}$, respectively.}
It is worth \qs{noting} that Taylor expansion formulation represents the relation between $\emph{ f }\left( x \right) $ and the Taylor series \qs{$S(x_p)$} at the expansion point $x_p$, 
\qs{and the scale of Taylor series would grow exponentially with the increase of derivative orders.}
\qs{As a result, the computational cost would also exhibit exponential growth.}
Although \cite{zhaotaylornet} proposes to decompose the Taylor series for high order derivatives to reduce \qs{the scale of expansion coefficients,} 
\qs{it is still computationally expensive for NeRF.}
\qs{Moreover, adopting higher-order Taylor expansion will lead to a rapidly increasing computational cost for each spatial point. Therefore, we advocate exploring the low-order Taylor expansion to design our efficient and effective representation.}

\par 
\qs{Specifically, we use a dense grid $G$ of size $w \times d \times h$ to represent the geometry function $f$,}
\qs{and make each vertex of voxels store the Taylor series containing the low-order derivatives of the objective function. We name this dense grid as \textbf{Taylor grid}, $\boldsymbol{V}_T$.}
Given \qs{an} input point $x$, 
\qs{we first query the Taylor voxel $\boldsymbol{v}_T$ it belongs to in the Taylor grid:}
\begin{equation}
    \boldsymbol{v}_T = {query}(x,\boldsymbol{V}_T),
\end{equation}
\qs{where $query(\cdot,\cdot)$ represents query operation.}  
\qs{Then, for each vertex $v_i (i=0\dots7)$ of the Taylor voxel, we can compute an approximate value $\tilde{f}(x;v_i)$ for the geometry function $f(x)$ based on its stored Taylor coefficients and expansion formation, that is:}
\begin{equation}
    \tilde{f}(x;v_i) = T'(x-v_i, S(v_i)),
\end{equation}
\qs{where $T'(x,v_i)$ is the 2-order version of Equation~\ref{Taylor_expansion}}.
With \qs{these} eight vertices and approximate values,  we convert the Taylor voxel into a linear voxel $\boldsymbol{v}_L$.
In this way, $f(x)$ is computed as:
\begin{equation} \label{taylor voxel}
\begin{split}
    \emph{f} \left(x\right) &=  interp\left(x,\boldsymbol{v}_\emph{L}\right)
\end{split}
\end{equation}
where $interp(\cdot,\cdot)$ represents trilinear interpolation. \qs{We term our representation as TaylorGrid.}

\qs{Based on the above implementation, we see that our proposed representation has following benefits:}
$\left(1\right)$ \qs{\emph{Representation form.} Our method is very lightweight. In contrast to SMLP, it does not require a large scale feature voxel or a shallow MLP to decode information. Moreover, it is implemented with query and interpolation operations. As a result, it has a better memory footprint and a faster converge speed.}
$\left(2\right)$ \qs{\emph{Representation power.} In comparison to Linear grid methods, our method introduces more non-linearity based on low-order Taylor expansion, thus enhancing representation power. This allows our method to achieve higher quality geometry modeling.} 
$\left(3\right)$ \qs{\emph{Application.} Since our method is simple and general, it can be easily plugged into any methods requiring implicit geometry representation. Next, we will introduce the application of TaylorGrid to geometry reconstruction and neural radiance fields respectively.}

\subsection{Geometry Reconstruction} \label{gemotry}

SDFs are functions  $\emph{ f}_{SDF} : \mathbb{R}^{3} \rightarrow \mathbb{R}$, 
\qs{representing} the shortest singed distance from a point $x$ in 3D space to a surface $\mathcal{S} = \partial \mathcal{M}$ of volume $\mathcal{M} \subset \mathbb{R}^{3}$.  
\qs{For a 3D point $x$, if its SDF value $f_{SDF}(x)$ is negative, it indicates that this point is inside of the surface $\mathcal{S}$. A positive SDF value indicates that this point is outside of the surface.}
\qs{In this way, the surface }$\mathcal{S}$ is implicitly represented as the zero level-set of $\emph{ f}_{SDF} $:
\begin{equation} 
\begin{split}
    \mathcal{S} = \left\{ x \in \mathbb{R}^{3} \vert \emph{ f}_{SDF} \left( x \right)= 0 \right \}.
\end{split}
\end{equation}
\qs{Here, we use our proposed TaylorGrid to model the $f_{SDF}$.}

\par \noindent \textbf{Optimization.} 
\qs{To} make the training process \qs{pay} more attention to the shape surface and speed up the training, we \qs{design} a reconstruction loss as follows:
\begin{equation} 
\begin{split}
    &\mathcal{L}_{recon} = \frac{1}{|\mathcal{P}|}
\sum_{x_i \in \mathcal{P}} \emph{w} (x_i) \;\Vert\hat{d}(x_i)- d(x_i)\Vert_2, 
\end{split}
\end{equation}
\qs{where $\mathcal{P}$ denotes the set of sampled points in the minibatch,} $\hat{d}(x_i) $ is the output SDF of model for input $x_i$, \qs{and} $d(x_i)$ is the \qs{ground} truth SDF. The $\emph{w}(x_i)$ is an adaptive weight 
and is computed as:
\begin{equation} 
\begin{split}
    &\emph{w} (x_i) = \max(\;w'(\,\hat{d}(x_i)\,),\;w'(\,d(x_i)\,)\;),  \\
    &\emph{w}' (d) =  1 - \lvert 2 \cdot \text{sigmoid}(k\cdot d) - 1 \rvert. 
\end{split}
\end{equation}
The $w(x_i)$ tends to $1$ when the $d(x_i)$ or $\hat{d}(x_i)$ is near 0, and tends to $0$ otherwise. This makes \qs{the optimization} \qs{pay} more attention to the surface of objects for both ground truth and predicted one. 
\qs{In addition}, directly optimizing the grid\qs{-based representation} with the reconstruction loss usually \qs{leads to} some noise \cite{fridovich2022plenoxels,max1995optical}. 
\qs{To alleviate this problem, we apply the total variation loss 
$\mathcal{L}_{TV}$ from Plenoxel \cite{fridovich2022plenoxels} and the Eikonal constraint $\mathcal{L}_{eik} $ used in \cite{IGR} to suppress the noise. In this way,}
our total loss is as follows:
\begin{equation} 
\begin{split}
    &\mathcal{L} = \mathcal{L}_{recon} + \lambda_1\mathcal{L}_{eik} + \lambda_2\mathcal{L}_{TV},
\end{split}
\end{equation}
\qs{where $\lambda_1$ and $\lambda_2$ are weight coefficients to balance different losses.}

\subsection{Neural Radiance Fields} \label{nerf_section}
To represent a 3D scene for novel view synthesis, Neural Radiance Fields (NeRFs) \cite{BenMildenhall2020NeRFRS} employ multi-layer perceptron (MLP) networks to map a 3D position $x$ and a viewing direction $v \left( \theta, \phi\right)$ to the corresponding density $\sigma$ and view-dependent color emission $c$. 
Specifically, NeRFs get the density $\sigma$ with $\emph{f}_\sigma : \mathbb{R}^3 \to \mathbb{R}$, and the  color emission $c$ with $\emph{f}_c : \mathbb{R}^3\cdot\mathbb{R}^2 \to \mathbb{R}^3$. The $\emph{f}_\sigma$ would map a 3D position $x \in \mathbb{R}^3 $ to  $\sigma \in \mathbb{R} $ while the  $\emph{f}_c$ would map the 3D position  $x \in \mathbb{R}^3 $ and the view direction $v  \in \mathbb{R}^2 $ to  $c \in \mathbb{R}^3 $. We formulate them with:
\begin{equation} 
\begin{split}
    &\sigma = \emph{f}_\sigma\left(x\right), \\
    &c = \emph{f}_c \left(x,v\right).
\end{split}
\end{equation}
Because the density $\sigma$ just depends on the 3D position $x$, which represents the geometry of the scene \qs{like} $\emph{f}_{SDF}\left(x\right)$, we can model the density $\sigma$ \qs{by using the TaylorGrid to} represent the $\emph{f}_\sigma\left(x\right)$. 
\qs{As for the color field, considering view-dependent appearance, we employ the spherical harmonic illumination used in \cite{fridovich2022plenoxels} or the hybrid representation with feature voxels used in \cite{voxGo} to represent the $\emph{f}_c\left(x,v\right)$.}
\par \noindent \textbf{Optimization.} 
\qs{For a rendered ray $r = o + t \cdot v$ cast from the camera origin $o$ along the viewing direction $v$, according to the optical model provided by Max \cite{max1995optical}, its corresponding rendered color $\hat{C}(r)$ is computed by accumulating the colors of sample points along the ray $r$:}
\qs{
\begin{equation}
\begin{split}
    \hat{C}(r) = \sum_{i=1}^{N} T_i (1 - \text{exp}(-\sigma_i \delta_i))c_i, \\
    \text{with} \; T_i = \text{exp}(-\sum_{j=1}^{i-1} \sigma_j \delta_j), 
\end{split}
\end{equation}
}
\qs{where $N$ is the total number of sample points along the ray and $\delta_i$ is the distance between adjacent sample points. Then, we optimize the Taylor series of TaylorGrid by minimizing the photo-metric loss between the predicted rendered color $\hat{C} \left(r\right)$ and the observed color $C\left(r\right)$}:
\begin{equation} 
\begin{split}
    \mathcal{L}_{photo}
    &= \frac{1}{\qs{|\mathcal{R}|}} \sum_{\qs{r \in \mathcal{R}}}\Vert\hat{C} \left(r\right) - C\left(r\right)\Vert^2_2,
\end{split}
\end{equation}%
where $\mathcal{R}$ \qs{denotes the set of rays in the minibatch}.

\section{Experiment}

\subsection{3D Geometry Reconstruction} \label{3D reconstruction}
\par  \noindent  \textbf{Dataset and Baselines.} In this study, we conduct 3D reconstruction experiments on the Stanford 3D models, including \emph{Armadillo}, \emph{Buddha}, \emph{Lucy}, \emph{Dragon}, \emph{ThaiStatue}, \emph{Bunny}, and \emph{Bimba}. 
\qs{In addition, we select} seven objects from Thingi10K \cite{Thingi10K} and Synthetic Indoor Scene \cite{mescheder2019occupancy} datasets for single shape reconstruction. 
We compare our approach with DeepSDF, feature grid combined with shallow MLP (SMLP) and linear grid (Linear). Moreover, we follow the Relu filed \cite{karnewar2022relu} that uses \qs{an} activation function, Tanh to introduce non-linearity to linear grid, \emph{i.e.}, linear grid+Tanh (LTanh). 
We use uniform sampling, near-surface sampling, and ray tracing-based methods to get sample points and their SDF for supervision, with a sampling ratio of 1:2:2 and total point number $500,000$ for all datasets.

\begin{table}[b]
  \centering
 \begin{tabular}{lcccccccc}
 \hline
   & \multicolumn{2}{c}{Standford model} &   & \multicolumn{2}{c}{Thingi10K} &   & \multicolumn{2}{c}{Scenes} \\
\cline{2-3}\cline{5-6}\cline{8-9}   & IoU↑ & CD↓ &   & IoU↑ & CD↓ &   & IoU↑ & CD↓ \\
\hline
 DeepSDF & 98.45  & 0.1141  &   & 99.24  & 0.1311  &   & 95.80  & 0.1398  \\
 \hline
 Linear$_{64}$ & 97.03  & 0.1218  &   & 99.10  & 0.1315  &   & 70.31  & 0.1726  \\
 LTanh$_{64}$ & 97.36  & 0.1230  &   & 99.17  & 0.1331  &   & 72.17  & 0.1735  \\
 SMLP$_{64}$ & \textbf{99.23}  & \textbf{0.1102}  &   & 99.59  & 0.1319  &   & \textbf{96.74}  & 0.1389  \\
 $1^{st}$ Taylor$_{64}$ & 98.87  & 0.1114  &   & 99.60  & 0.1273  &   & 93.47  & 0.1398  \\
 Taylor$_{64}$ & 99.09  & 0.1104  &   & \textbf{99.69}  & \textbf{0.1273}  &   & 95.35  & \textbf{0.1387}  \\
 \hline
 Linear$_{128}$ & 98.76  & 0.1109  &   & 99.61  & 0.1272  &   & 90.54  & 0.1404  \\
 LTanh$_{128}$ & 99.07  & 0.1114  &   & 99.79  & 0.1303  &   & 92.04  & 0.1416  \\
 SMLP$_{128}$ & 99.44  & 0.1102  &   & 99.66  & 0.1289  &   & 97.01  & 0.1383  \\
 $1^{st}$ Taylor$_{128}$ & \textbf{99.59 } & \textbf{0.1096}  &   & 99.52  & 0.1298  &   & 97.43  & 0.1376  \\
 Taylor$_{128}$  & 99.56  & \textbf{0.1096}  &   & \textbf{99.81 } & \textbf{0.1268 } &   & \textbf{97.45 } & \textbf{0.1374 } \\
 \hline
 \end{tabular}%
    \caption{Quantitative evaluation on the Stanford scans models, Thingi10K and Synthetic Indoor Scenes.} 
 \label{3D_result_table}
\end{table}%
\par \noindent  \textbf{Implementation Details and Evaluation Metrics.} \qs{Our proposed TaylorGrid is implemented in CUDA}. We adopt the PyTorch \cite{paszke2019pytorch} framework to train and test it. We use a progressive training approach \cite{karnewar2022relu} with a fixed learning rate of 0.003, and set the $\lambda_1$ to 0.0001 ,$\lambda_2$ to 0.00002. 
For the SMLP method, we use a neural voxel with 16 feature channels and one-layer MLP with 128 hidden units. 
For the Linear method,  we \qs{also use} $\mathcal{L}_{eik}$ and $\mathcal{L}_{TV}$ to improve the reconstruction quality. We adopt two resolution settings, $64^3$ and $128^3$, to test grid-based methods. 
We evaluate the reconstruction results using the following metrics: intersection-over-union (IoU) and chamfer-distance-L1 (CD). 
We calculate IoU by uniformly sampling $100k$ points and using the \qs{evaluation} code offered by Occupancy-Net \cite{mescheder2019occupancy}. After extracting the mesh with marching cube \cite{marchcube}, we calculate the CD with the code provided by PyTorch3D \cite{ravi2020pytorch3d}. 
All experiments are performed on a single NVIDIA A10 GPU with 24G memory.

\begin{figure*}  [!h] 
\centering
  \includegraphics[width=0.98\linewidth]{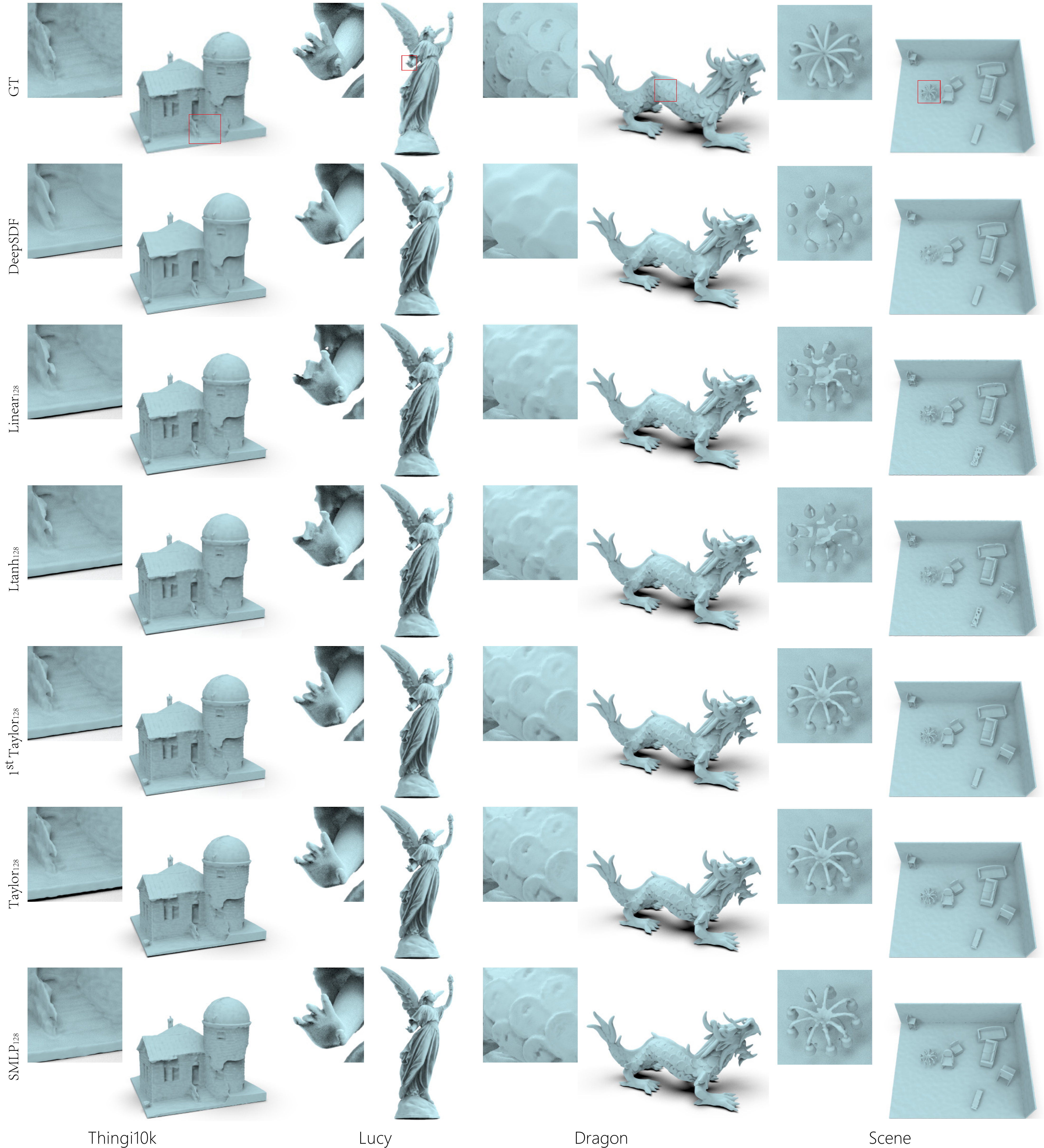}
  \caption{Qualitative results for the geometry reconstruction. Comparisons are made with DeepSDF, linear grid, shallow MLP, and Taylor.
  For Taylor, a grid of $128^3$ resolution is used.
  }
  \label{3D_result_fig}
\end{figure*}

\begin{figure}[t]
\centering
  \includegraphics[width=0.7\linewidth]{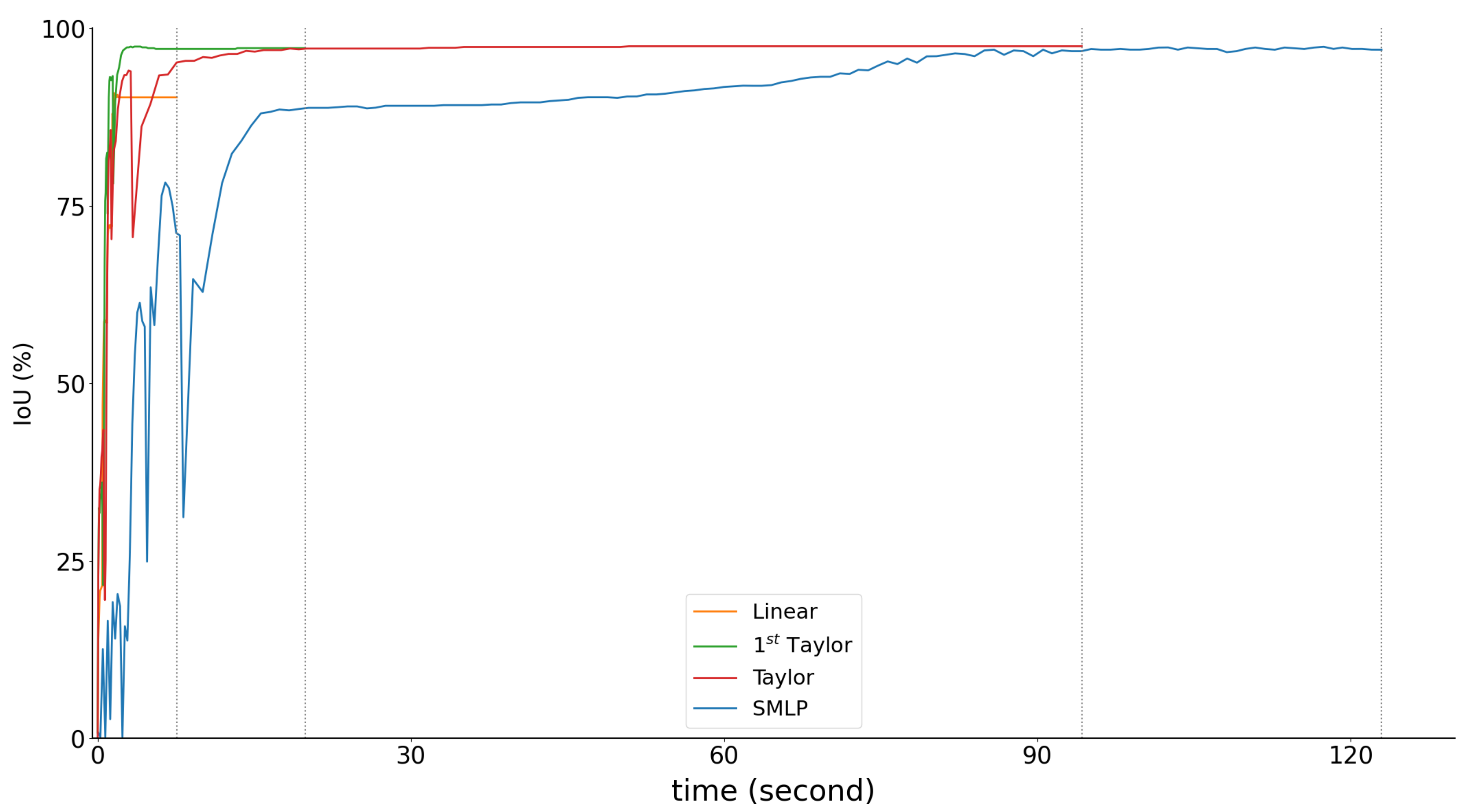} 
  \caption{The chart shown the IoU curve over training time of Linear, SMLP, $1^{st}$ Taylor and Taylor. We let all of methods have the same training epochs, the gray dash vertical lines indicate the end of training. For the same training epochs,  our 
 method, especially $1^{st}$ Taylor,  can finish training more quickly. Our method can convergence as fast as Linear, and has a good quality as SMLP.
  }
  \label{time_3D}
\end{figure}

\par  \noindent  \textbf{Comparison Results.} We present the comparison results of TaylorGrid with DeepSDF, Linear and SMLP. 
In the Table~\ref{3D_result_table}, we provide quantitative evaluation results on Stanford models, Thingi10K and Synthetic Indoor Scene Dataset. 
\qs{As can be seen, in the case of a resolution of $128^3$, our method achieves the best CD results and almost the best IoU results on all datasets. In the case of a resolution of $64^3$, our method is very competitive with SMLP. This demonstrates the superiority of our method in geometry reconstruction.} 
The  qualitative results \qs{at a resolution of} $128^3$ are shown in Fig.~\ref{3D_result_fig}.
\qs{Although linear grid methods can well fit simple data, such as Thingi10K, they cannot recover details and reconstruct large scenes due to their limited representation power.} 
\qs{In contrast, thanks to our proposed TaylorGrid, our method can reconstruct details well, such as the fingers in the Stanford LUCY model, which are difficult for comparison methods.}

\begin{table}[t]
  \centering

 \begin{tabular}{c|c|ccccc}
 \hline
 \multicolumn{1}{r}{} &   & Linear & $1^{st}$ Taylor & Taylor & SMLP \\
 \hline
 \multirow{3}[2]{*}{\rotatebox{90}{Scene}} & Time (s)\textdownarrow & 7.0 & 27.1 & 95.0 & 121.4\\
   & Model Size (MB)\textdownarrow & 9.2  & 32.7  & 106.0  & 147.0 \\
   & Quality (IoU)\textuparrow & 90.54  & 97.43  & 97.45  & 97.01\\
 \hline
 \multirow{3}[2]{*}{\rotatebox{90}{\footnotesize NeRF-SH}} & Time (s)\textdownarrow & 278.0  & 360.0  & 454.0  & 1149.0\\
   & Model Size (GB)\textdownarrow & 0.683  & 0.756  & 0.781  & 0.980  &\\
   & Quality (PSNR)\textuparrow & 30.18  & 30.95  & 31.08  & 31.27  &\\
 \hline
 \end{tabular}%
 \caption{Comparison results for training time, model size and  quality of different methods, on the Scene geometry reconstruction and  neural radiance fields with spherical harmonics.} 
 \label{3D_Q_M_table}
\end{table}

\qs{We compare the efficiency and model size of different methods in Table~\ref{3D_Q_M_table} and Fig.~\ref{time_3D}. Our method converges quickly, only taking about 2 minutes to reach the best. In contrast, the DeepSDF takes about 25 minutes and the SMLP takes about 8 minutes. Although the Linear only takes 7 seconds to converge, its reconstruction quality is unsatisfactory. In terms of model size, since the Linear only considers the zero-order Taylor series, thus it has the least number of parameters. While SMLP still requires a shallow MLP, it has the largest number of parameters. Our method does not require shallow MLPs but introduces low-order Taylor series. Therefore, the parameter size of our method is between these two parameter sizes.} 
\par \noindent \textbf{Analysis.} \qs{We first investigate the effect of the order of Taylor expansion, including zero-order, one-order and second-order. These three settings correspond to the Linear, $1^{st}$ Taylor and Taylor in Table~\ref{3D_result_table}, respectively. We observe that, as the order increases, the performance continues to improve, and the performance is basically saturated with the second-order. Considering the time cost and model size, we think that the second-order of Taylor expansion can meet most our needs here.} 
\qs{In terms of grid resolutions, increasing grid resolutions will improve all grid-based methods, including ours.}
\par
\begin{figure*}[t] 
  \includegraphics[width=\linewidth]{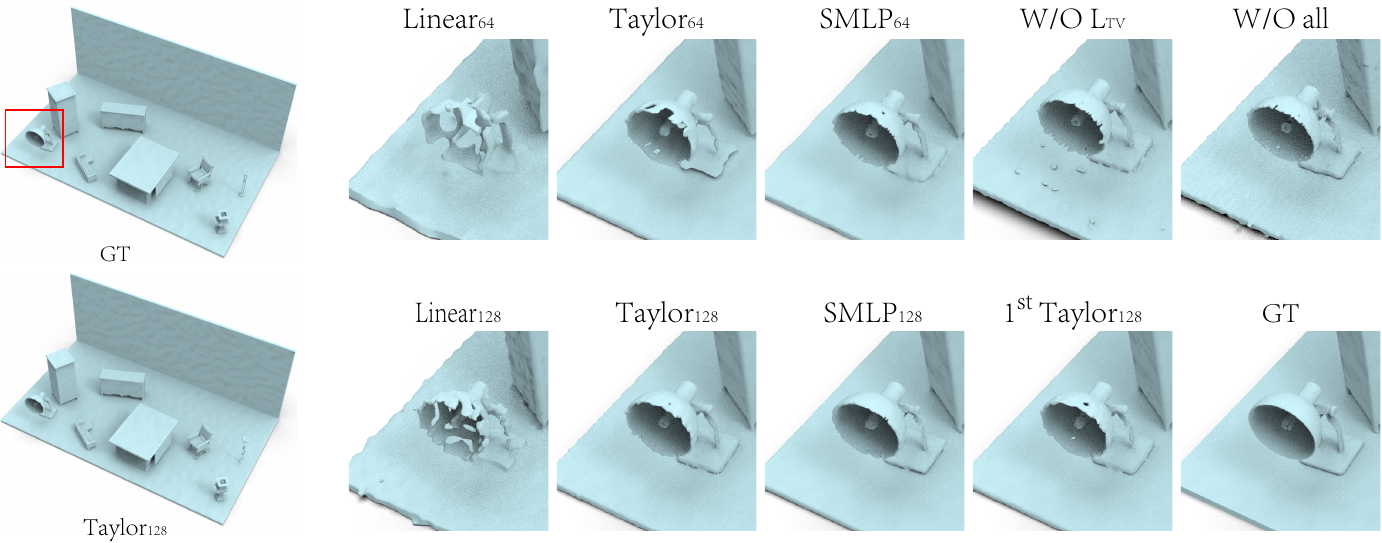}
  \caption{Qualitative results for ablation study on geometry reconstruction. Comparisons are made to evaluate the effects for resolution with 64 and 128, shallow MLP, loss of $L_{TV}$ and first order Taylor. Among the results, the full Taylor$_{128}$ model achieves the best quality.
  }
  \label{3D_ablation}
\end{figure*}

 \noindent \textbf{Ablation Study.} \qs{We conduct an ablation study to validate the effectiveness of the total variation loss $\mathcal{L}_{TV}$ and weighting scheme.} $\mathcal{L}_{TV}$ is proposed by Plenoxels \cite{fridovich2022plenoxels} for neural radiance fields, 
\qs{while we explore it for geometry reconstruction. By adding different components step by step, we obtain the ablation results in Table~\ref{3D_ab_table}. We see that paying more attention to surface optimization helps to achieve more accurate reconstruction. Although we have applied the eikonal constraint to regularize the geometry, the grid-based methods are still sensitive to noise. After adding the total variation loss, our method surpasses the noise better, thus achieving reconstruction results with higher quality, as shown in Fig.~\ref{3D_ablation}.} 

\begin{table}[t]
  \centering
\begin{tabular}{ccccrccrcc}
\hline
    \multirow{2}{*}{weighting} & \multirow{2}{*}{$\mathcal{L}_{TV}$} & \multicolumn{2}{c}{Standford model} &   & \multicolumn{2}{c}{Thingi10K} &   & \multicolumn{2}{c}{Scenes} \\
   \cline{3-4}\cline{6-7}\cline{9-10}  & & IoU↑ & CD↓ &   & IoU↑ & CD↓ &   & IoU↑ & CD↓ \\
   \hline
   & & 99.65  & 0.1135  &   & 99.75  & 0.1282  &   & 95.94  & 0.1363  \\
  \checkmark &  & \textbf{99.69 } & \textbf{0.1096}  &   & \textbf{99.82 } & 0.1273  &   & 96.51  & 0.1417  \\
 \checkmark & \checkmark & 99.56  & \textbf{0.1096}  &   & 99.81  & \textbf{0.1268 } &   & \textbf{97.45 } & \textbf{0.1374 } \\
 \hline
 \end{tabular} 
 \caption{Ablation study on the $\mathcal{L}_{TV}$ and the weighting scheme.} 
 \label{3D_ab_table}
\end{table}

\subsection{Neural Radiance Fields}
\par  \noindent  \textbf{Datasets and Baselines.} We use Synthetic-NeRF dataset to evaluate the neural radiance fields performance of different methods. 
We set the image resolution to $800 \times 800$. 
100 images are used for training and 200 images for testing per object. 
\qs{As mentioned before, neural radiance fields are affected by view-dependent effects. Therefore, we only use our proposed TaylorGrid to represent the density field, and adopt the spherical harmonic coefficients used in Plenoxels \cite{fridovich2022plenoxels} and the hybrid representation used in DVGO \cite{voxGo} to represent the color field respectively.}

\qs{We compare our geometry representation with the following methods: linear grid (Linear), feature grid combined with shallow MLP (SMLP) and the Linear grid with a Softplus activation function (LS). }
It is worth noting that LS is the \qs{original density representation} of DVGO.

\par  \noindent  \textbf{Implementation Details and Evaluation Metrics.} 
\qs{For a fair comparison,}
we \qs{train} all methods in a same way. 
In fact, no matter what color representation is combine with any geometry representation, 
we use the coarse-to-fine training strategy in DVGO. 
\qs{That is,} we first use the linear grid for coarse geometry searching, 
then we focus on a smaller subspace to reconstruct the geometry details and view-dependent effects. 
Please refer to the DVGO for more details. 
Similar to the 3D geometry reconstruction, We adopt two resolution settings, $128^3$ and $256^3$, to test grid-based methods
We adopt the widely used PSNR and LPIPS to evaluate  novel view synthesis results, as well as presenting the average training time. 
All experiments are conducted using PyTorch on a single A10 GPU with 24G memory.

\begin{table*}[!t]
    \centering
    \resizebox{\linewidth}{!}{
    \setlength{\tabcolsep}{0.2pt}
    \begin{tabular}{cl|cccccccccccccccc|ccc}
        \hline
         & \multirow{2}{*}{Method} & \multicolumn{2}{c}{CHAIR} & \multicolumn{2}{c}{DRUMS} & \multicolumn{2}{c}{FICUS} & \multicolumn{2}{c}{HOTDOG} & \multicolumn{2}{c}{LEGO} & \multicolumn{2}{c}{MATERIALS} & \multicolumn{2}{c}{MIC} & \multicolumn{2}{c|}{SHIP} & \multicolumn{3}{c}{Avg.} \\
         \cline{3-21}
         & & PSNR & LPIPS & PSNR & LPIPS & PSNR & LPIPS & PSNR & LPIPS & PSNR & LPIPS & PSNR & LPIPS & PSNR & LPIPS & PSNR & LPIPS & PSNR & LPIPS & Time \\
         \hline
         \multirow{10}{7pt}{\rotatebox{90}{NeRF-SH}} & \multicolumn{1}{l|}{Linear$_{128}$} & 31.85  & 0.052 & 24.53  & 0.104  & 30.74  & 0.037  & 35.12  & 0.054  & 31.70  & 0.060  & 29.04  & 0.067  & 30.96  & 0.037  & 27.50  & 0.196  & 30.18  & 0.076  & 0:04:38 \\
         & \multicolumn{1}{l|}{LS$_{128}$} & 31.75  & 0.052 & 24.53  & 0.104  & 30.62  & 0.038  & 35.12  & 0.054  & 31.70  & 0.060  & 29.04  & 0.067  & 31.09  & 0.035  & 27.42  & 0.198  & 30.16  & 0.08  & 0:05:36 \\
         & \multicolumn{1}{l|}{$1^{st}$ Taylor$_{128}$} & 32.51  & 0.044  & 24.93  & 0.094  & 32.09  & 0.028  & 35.91  & 0.046  & 33.02  & 0.046  & 29.25  & 0.060  & 31.58  & 0.028  & 28.32  & 0.180  & 30.95  & 0.066  & 0:06:00 \\
         & \multicolumn{1}{l|}{Taylor$_{128}$} & 32.64  & 0.041  & 25.06  & 0.088  & 31.89  & 0.029  & 35.82  & 0.047  & 33.68  & 0.035  & 29.42  & 0.060  & 31.64  & 0.029  & 28.47  & 0.175  & 31.08  & 0.063  & 0:07:34 \\
         & \multicolumn{1}{l|}{SMLP$_{128}$} & 33.17  & 0.037  & 25.09  & 0.091  & 32.54  & 0.028  & 35.96  & 0.049  & 33.29  & 0.042  & 29.47  & 0.057  & 32.15  & 0.025  & 28.54  & 0.175  & 31.27  & 0.063  & 0:19:09 \\
         \cline{2-21}
         & \multicolumn{1}{l|}{Linear$_{256}$} & 34.34  & 0.026  & 25.30  & 0.078  & 32.40  & 0.026  & 36.39  & \textbf{0.038}  & 34.13  & 0.030  & 29.28  & 0.055  & 33.34  & 0.017  & 29.25  & 0.150  & 31.80  & 0.052  & 0:10:37 \\
         & \multicolumn{1}{l|}{LS$_{256}$} & 34.23  & 0.026  & 25.26  & 0.077  & 32.03  & 0.029  & 36.32  & 0.039  & 33.95  & 0.030  & 29.02  & 0.059  & 33.33  & 0.017  & 28.99  & 0.157  & 31.64  & 0.054  & 0:11:16 \\
         & \multicolumn{1}{l|}{$1^{st}$ Taylor$_{256}$} & \textbf{34.78 } & \textbf{0.023 } & 25.43  & 0.076  & 32.74  & \textbf{0.025 } & 36.55  & \textbf{0.038}  & 34.52  & 0.027 & 29.26  & \textbf{0.054 } & 34.00  & \textbf{0.014 } & \textbf{29.89 } & \textbf{0.142 } & 32.15  & \textbf{0.050 } & 0:13:08 \\
         & \multicolumn{1}{l|}{Taylor$_{256}$} & 34.67  & 0.024  & \textbf{25.50 } & \textbf{0.075 } & 32.61  & 0.028  & \textbf{36.56 } & 0.039  & \textbf{35.02 } & \textbf{0.024 } & \textbf{29.39 } & 0.057  & \textbf{34.51 } & 0.016  & 29.71  & 0.143  & \textbf{32.25 } & 0.051  & 0:16:24 \\
         & \multicolumn{1}{l|}{SMLP$_{256}$} & 34.75  & 0.026  & 25.39  & 0.078  & \textbf{32.75 } & 0.028  & 35.84  & 0.047  & 34.12  & 0.030  & 29.15  & 0.057  & 34.06  & 0.018  & 29.59  & 0.147  & 31.95  & 0.054  & 0:35:10 \\
         \hline
         \multirow{10}{7pt}{\rotatebox{90}{DVGO}} & \multicolumn{1}{l|}{Linear$_{128}$} & 33.36  & 0.034 & 25.12  & 0.089  & 31.93  & 0.029  & 36.43  & 0.038  & 33.42  & 0.035  & 29.38  & 0.063  & 32.42  & 0.024  & 28.47  & 0.175  & 31.32  & 0.061  & 0:04:21 \\
         & \multicolumn{1}{l|}{LS$_{128}$} & 33.22  & 0.034  & 25.11  & 0.088  & 31.83  & 0.029  & 36.41  & 0.038  & 33.34  & 0.036  & 29.33  & 0.064  & 32.41  & 0.024  & 28.40  & 0.177  & 31.26  & 0.061  & 0:05:23 \\
         & \multicolumn{1}{l|}{$1^{st}$ Taylor$_{128}$} & 33.38  & 0.034  & 25.53  & 0.080  & 33.42  & 0.021  & 36.65 & 0.036  &34.92  & 0.028  & 29.53  & 0.058  & 32.54  & 0.021  & 29.11  & 0.164  & 31.88  & 0.055  & 0:04:48 \\
         & \multicolumn{1}{l|}{Taylor$_{128}$} & 33.42  & 0.034  & 25.56  & 0.078  & 33.45  & 0.021  & 36.62  & 0.037  & 35.09  & 0.028  & 29.53  & 0.057  & 32.57  & 0.021  & 29.44  & 0.161  & 31.96  & 0.055  & 0:07:35 \\
         & \multicolumn{1}{l|}{SMLP$_{128}$} & 33.54  & 0.032  & 25.59  & 0.078  & 33.97  & 0.021  & 36.36  & 0.040  & 34.53  & 0.029  & 29.72  & 0.055  & 32.77  & 0.021  & 29.23  & 0.163  & 31.96  & 0.055  & 0:16:23 \\
         \cline{2-21}
         & \multicolumn{1}{l|}{Linear$_{256}$} & 35.40  & \textbf{0.019}  & 25.79  & 0.070  & 33.57  & 0.021  & 36.98  & \textbf{0.030}  & 35.89  & 0.020  & 29.60  & 0.055  & 34.69  & 0.012  & 30.01  & 0.140  & 32.74  & 0.046  & 0:07:45 \\
         & \multicolumn{1}{l|}{LS$_{256}$} & 35.25  & 0.020  & 25.57  & 0.072  & 33.25  & 0.022  & 37.04  & 0.031  & 35.62  & 0.022  & 29.24  & 0.061  & 34.46  & 0.012  & 29.75  & 0.143  & 32.52  & 0.048  & 0:11:26 \\
         & \multicolumn{1}{r|}{$1^{st}$ Taylor$_{256}$} & 35.31  & 0.020  & 25.96  & 0.066  & 34.15  & 0.019  & \textbf{37.07 } & \textbf{0.030}  & 36.25  & \textbf{0.019}  & 29.58  & 0.054  & \textbf{35.13 } & \textbf{0.011}  & 30.54  & \textbf{0.134}  & \textbf{33.00 } & \textbf{0.044}  & 0:10:16 \\
         & \multicolumn{1}{l|}{Taylor$_{256}$} & 35.33  & 0.020  & 25.95  & 0.065  & 33.99  & 0.019  & 37.05  & 0.031  & \textbf{36.26 } & \textbf{0.019}  & 29.57  & 0.054  & 35.10  & \textbf{0.011}  & \textbf{30.65 } & \textbf{0.134}  & 32.99  & \textbf{0.044}  & 0:16:10 \\
         & \multicolumn{1}{l|}{SMLP$_{256}$} & \textbf{35.44 } & \textbf{0.019}  & \textbf{26.00 } & \textbf{0.064 } & \textbf{34.57 } & \textbf{0.018 } & 36.77  & 0.031  & 35.60  & 0.021  & \textbf{29.63 } & \textbf{0.052 } & 34.50  & 0.012  & 30.38  & 0.136  & 32.86  & \textbf{0.044}  & 0:30:09 \\
         \hline
         \multicolumn{2}{c|}{NeRF} & 33.00  & 0.046  & 25.01  & 0.091  & 30.13  & 0.044  & 36.18  & 0.121  & 32.54  & 0.050  & 29.62  & 0.063  & 32.91  & 0.028  & 28.65  & 0.206 & 31.01  & 0.081  & - \\
        \hline
    \end{tabular}
    }
    \caption{Quantitative results of our method with linear methods and shallow MLP mehtods on the Synthetic NeRF scenes. For PSNR, higher is better. For LPIPS, lower is better.}
    \label{Nerf_show_table}
\end{table*}

\begin{figure}[!t]
  \centering
  \includegraphics[width=1.0\linewidth]{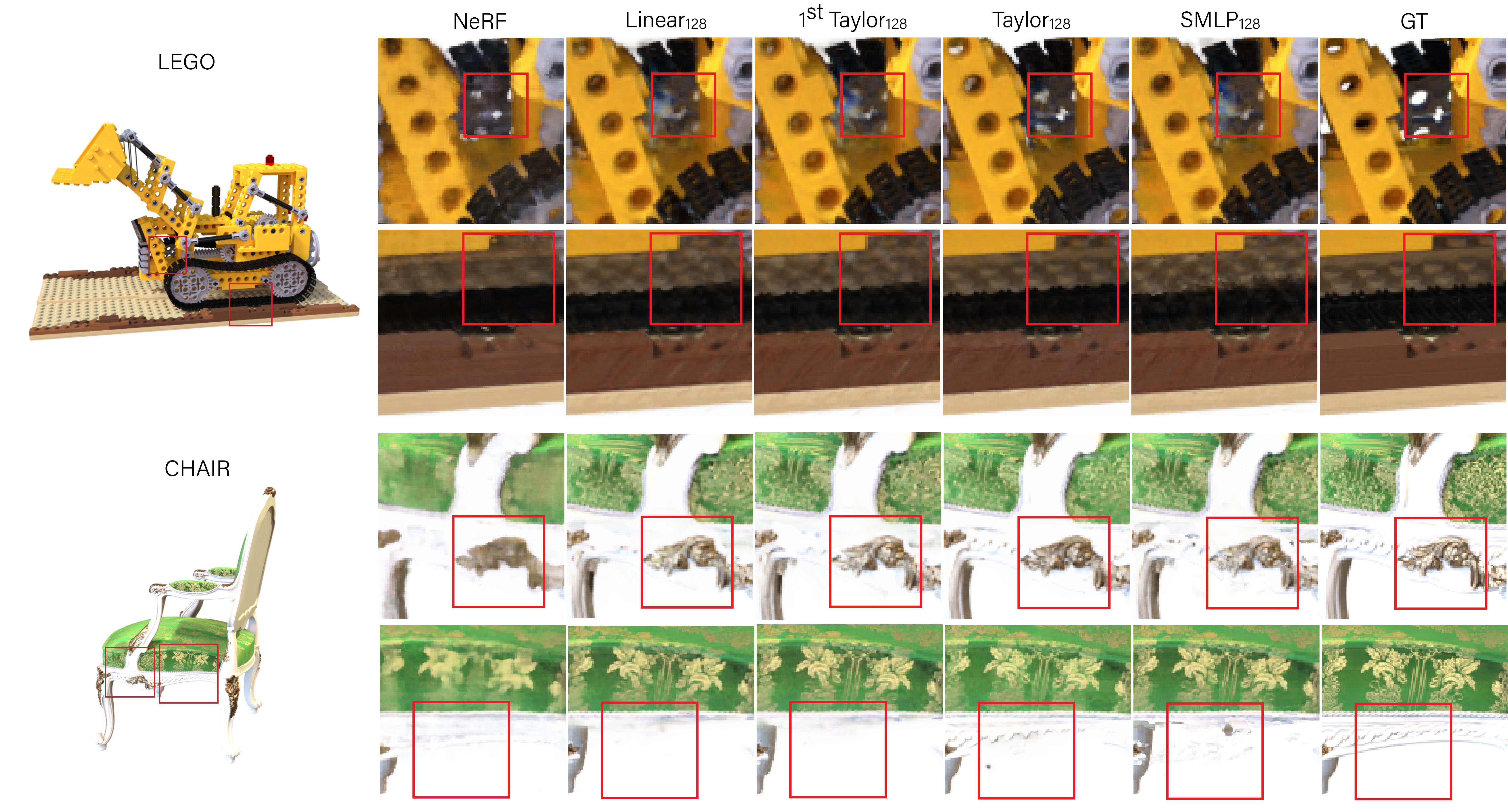}
  \caption{Qualitative results for novel view synthesis. Comparisons are made with linear grid, shallow MLP, heavy MLP (NeRF) and different orders of TaylorGrid on Synthetic-NeRF dataset.
  }
  \label{nerf_result_fig}
\end{figure}

\par  \noindent  \textbf{Comparison Results.} We present the qualitative results at a resolution of 128 in Fig.~\ref{nerf_result_fig}, and the quantitative comparison results in Table~\ref{Nerf_show_table}. \qs{We see that, by introducing non-linearity, grid-based methods almost achieve better PSNR and LPIPS than the original NeRF. Furthermore, as shown in the red boxes of Fig.~\ref{nerf_result_fig}, compared to the original NeRF that only uses MLPs to represent 3D scenes, grid-based methods are beneficial to model details. Although LS methods also introduce non-linearity by activation functions, their representation power is inferior to our Taylor-based representation. More importantly, our method achieves a better speed performance and a comparable quality result with SMLP. In particular, under the same settings, our method takes almost half the time of SMLP to achieve competitive results.} 
This is because both TaylorGrid and SMLP leverage powerful grids to model 3D scenes, but our introduced Taylor expansion is an explicit geometry constraint for feature grids like Linear methods.
\qs{Our method} can therefore converge as quickly as Linear, while achieving the competitive quality with SMLP. In addition, the results in Table~\ref{3D_Q_M_table} also show that our method is lightweight compared with SMLP thanks to our low-order Taylor series.

\begin{figure*} [!t] 
  \includegraphics[width=1.0\linewidth]{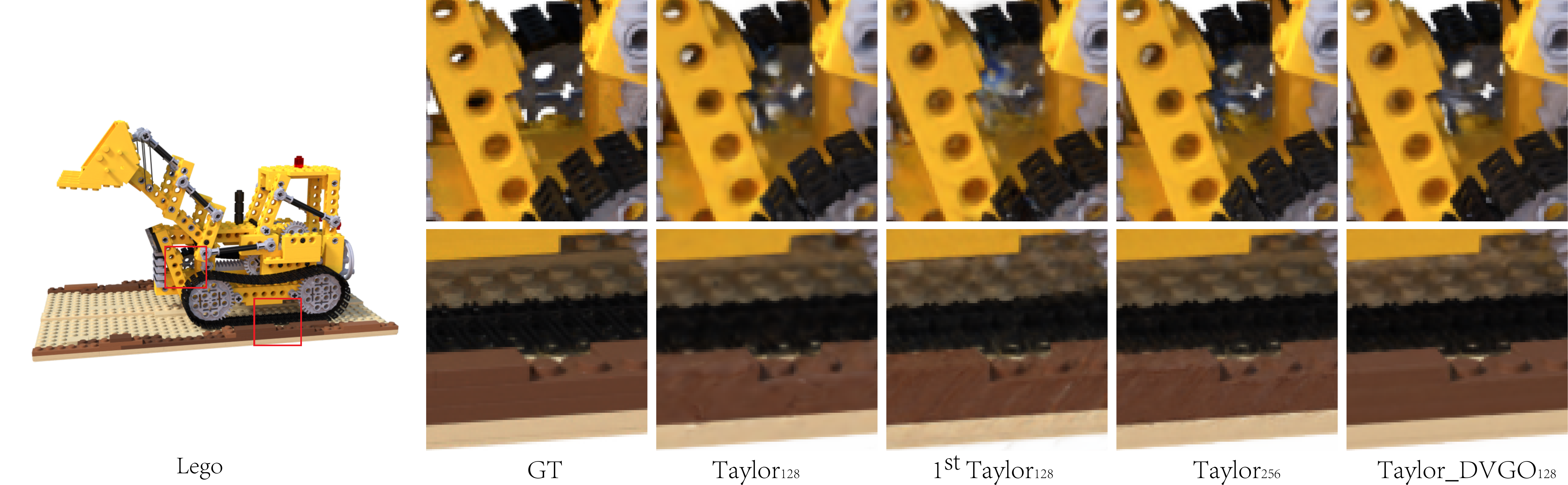}
  \caption{Qualitative results for NeRF ablation. Comparisons are made for applying TaylorGrid with different resolutions, order and NeRF backbones.
  }
   \label{nerf_abla_fig}
\end{figure*}

\par \noindent \textbf{Analysis.} \qs{Similar to geometry reconstruction, we further investigate the effects of Taylor expansion orders and grid resolutions. The results in Table~\ref{Nerf_show_table} basically show the same conclusion with the geometry reconstruction. As the order increases, the performance of novel view synthesis continues to improve and is almost saturated with the second-order (cf. Fig.~\ref{nerf_abla_fig}). Therefore, we advocate using low-order Taylor series to construct our method. We also see that increasing grid resolutions facilitates all grid-based methods. In addition, we observe that how to model the color field is also important to novel view synthesis. The results demonstrate that using the hybrid representation in DVGO to model the color field is better than the spherical harmonics in Plenoxels.}

\section{Conclusions}

\qs{In this paper, we have proposed TaylorGrid, a novel implicit field representation, which bridges the gap between linear grid and neural voxels by employing low-order Taylor expansion for direct grid optimization. In this way, our method achieves a better trade-off between efficiency and representation power, making a step further towards fast and high-quality implicit field learning.}
TaylorGrid can be easily plugged into other representation learning frameworks, replacing the linear grid or neural voxels. 
\qs{Extensive experiments on learning signed distance field and neural radiance field show that our method achieves superior performance on implicit field learning tasks. This demonstrates that our TaylorGrid has broad prospects in practice.}


On the other hand, there are still some limitations in our approach. 
First, our method also faces the disadvantage of all grid-based approaches, where memory consumption increases considerably with larger grid resolution. 
Second, to further improve the representation power of \qs{TaylorGrid}, one way is to \qs{increase} the order of Taylor expansion, but this will also increase the time overhead, hindering the practical applications of TaylorGrid. 
\qs{In the future, one direction is to explore the combination of the sparse data structure like Instant-NGP \cite{muller2022instant} or the grid decomposition scheme like TensoRF \cite{chen2022tensorf} with our Taylor-based method to further improve efficiency and representation power.}
In addition, although TaylorGrid is used to encode and predict scalar field values such as signed distance and density, it may be easily adapted to \qs{the tasks of predicting} non-scalar results, such as color. 
However, the Taylor series \qs{will become complicated} as the output dimension increases, which may \qs{lead to} undesirable speed issues. 
Last but not least, as Taylor expansion is continuous, it maybe difficult to express non-continuous information with the TaylorGrid. 
How to incorporate continuity and non-continuity into a uniform representation is another interesting future direction for implicit field learning.


\bibliography{mybibfile}
\end{document}